\documentclass[11pt,a4paper]{article}
\usepackage{times}
\usepackage{latexsym}
\usepackage{xcolor}
\usepackage{url}
\usepackage{adjustbox}
\usepackage{multirow}
\usepackage[hyperref]{naaclhlt2018}
\aclfinalcopy 
\title{Towards Inference-Oriented Reading Comprehension: ParallelQA}

\author{Soumya Wadhwa\thanks{\enskip Equal Contribution} \qquad
  Varsha Embar\footnotemark[1] \qquad
  Matthias Grabmair \qquad
  Eric Nyberg \\
  Carnegie Mellon University \\
  {\tt \{soumyaw, vembar, mgrabmai, en09\}@andrew.cmu.edu} \\
  } 

\date{}

\begin{document}
\maketitle
\begin{abstract}
In this paper, we investigate the tendency of end-to-end neural Machine Reading Comprehension (MRC) models to match shallow patterns rather than perform inference-oriented reasoning on RC benchmarks. We aim to test the ability of these systems to answer questions which focus on referential inference. We propose ParallelQA, a strategy to formulate such questions using parallel passages. We also demonstrate that existing neural models fail to generalize well to this setting. 

\end{abstract}

\section{Introduction}
Reading Comprehension (RC) is the task of reading a body of text and answering questions about it. It requires a deep understanding of the information presented in order to reason about entities, actions, events, and their interrelationships. This necessitates language understanding skills as well as the cognitive ability to draw inferences.

Recent efforts in creating large-scale datasets have triggered a renewed interest in the RC task, with subsequent development of complex end-to-end solutions featuring neural models. While these models do exceedingly well on the specific datasets they are developed for (some reaching or even surpassing human performance), they do not perform proportionally across datasets. \citet{weissenborn2017fastqa} have shown that using a context or type matching heuristic to derive simple neural baseline architectures can achieve comparable results. Our experiments also indicate that pattern matching can work well on these datasets.

Inference, an important RC skill \cite{ETS2:ETS200192, strange1980rc}, is the ability to understand the meaning of text without all the information being stated explicitly. Table \ref{inf-types}, Section \ref{sec:supplemental} describes the types of inference that we may encounter while comprehending a passage along with the cues that help perform such reasoning. Although state-of-the-art deep learning models for machine reading are believed to have such reasoning capabilities, the limited ability of these models to generalize indicates certain shortcomings. We believe that it is important to develop benchmarks which give a realistic sense of a system's RC capabilities. Thus, our goal in this paper is two-fold:

\noindent\textbf{Proof of Concept}: We propose a method to create an RC dataset that assesses a model's ability to:
\vspace{-0.8em}
\begin{itemize}
\setlength\itemsep{-0.4em}
\item move beyond lexical pattern matching between the question and passage,
\item infer the correct answers to questions which contain referring expressions, and
\item generalize to different language styles.
\end{itemize}
\vspace{-0.6em}
\noindent\textbf{Analysis of Existing Models}: We test three end-to-end neural MRC models, which perform well on SQuAD \cite{rajpurkar2016squad}, on a few question-answer pairs generated using our methodology. We demonstrate that it is indeed difficult for these systems to answer such questions, also indicating their tendencies to resort to shallow pattern matching and overfit to training data.

\section{Existing Datasets}
In this work, we focus on datasets with multi-word spans as answers rather than cloze-style RC datasets like MCTest \cite{richardson2013mctest}, CNN / Daily Mail \cite{hermann2015teaching} and Children's Book Test \cite{weston2015towards}. 

The Stanford Question Answering Dataset (SQuAD) \cite{rajpurkar2016squad} was one of the first large scale RC datasets (over 100k QA pairs), where the answer to each question is a span in the given passage. For its collection, different sets of crowd-workers were asked to formulate questions and answers using passages obtained from $\sim$500 Wikipedia articles. However, this resulted in the questions having similar word patterns to the sentences containing the answers. 
We empirically demonstrate this in Table \ref{statsexpts}, where we observed that the sentence in the passage with the highest lexical similarity to the question contained the answer $\sim$80\% of the time.
Final answers tend to be short, with an average span length of around 3 tokens, and are largely entities (40.88\%). \citet{subramanian2017neural} and \citet{yang2017semi} provide evidence for regular patterns in candidate answers that neural models can exploit. We show in subsequent sections that models which perform well on SQuAD rely on lexical pattern matching, and are also not robust to variance in language style.
\begin{table}[th!]
\begin{center}
\resizebox{\linewidth}{!}{
\begin{tabular}{|c|c|c|c|}
\hline \bf Metric & \bf SQuAD & \bf NewsQA & \bf ParallelQA\\ \hline
Jaccard & 79.28\% & 38.11\% & 27.45\% \\
TF-IDF & 81.32\% & 51.86\% & 31.37\% \\
BM25 & 74.26\% & 43.45\% & 27.45\% \\
\hline
\end{tabular}
}
\end{center}
\caption{Sentence Retrieval Performance using Jaccard similarity \cite{jaccard1912distribution}, TF-IDF overlap \cite{sparck1972statistical} and BM-25 overlap \cite{robertson1994okapi} scoring metrics}
\label{statsexpts}
\end{table}

To alleviate the lack of topic diversity in SQuAD, NewsQA \cite{trischler2016newsqa} was created from 12,744 news articles sampled from CNN/Daily Mail. To ensure lexical diversity, one set of crowd-workers generated questions using only an abstractive summary, while the answer spans were marked in the full article by another set of crowd-workers. However, news articles tend to encourage questions that point to entities, and the dataset does not specifically focus on inference. Determining the exact answer span is harder, but this may be due to the use of only news highlights to generate questions; this may induce noise in the answer spans marked in the news articles since the question might not be exactly apt.

To prevent annotation bias, SearchQA \cite{dunn2017searchqa} starts with question-answer pairs from Jeopardy! and adds documents retrieved by a search engine for each question as its context. However, the questions are mostly factoid. \citet{kovcisky2017narrativeqa} found that 80\% of answers are bigrams or unigrams, and 99\% contain 5 or fewer tokens, with many answers being named entities. TriviaQA \cite{joshi2017triviaqa} similarly includes question-answer pairs authored by trivia enthusiasts along with independently-gathered evidence documents which provide distant supervision for answering the questions.

These datasets have facilitated the development of new QA models, but we believe there are several important aspects of RC that remain untested.

\section{ParallelQA}
In an RC task, there is a need to incorporate questions that require not just lexical and syntactic prowess, but reference resolution, multiple steps of reasoning, and use of world knowledge. These capabilities ultimately lead to global rather than sentence-level understanding of text. The construction of a large-scale dataset of this nature is a challenging task. We take a small step in this direction by focusing on referential inference.\footnote{Referential inference is the process of identifying the discourse and/or real-world entity referred to by a linguistic expression (name, noun, pronoun, etc.).} WikiHop \cite{welbl2017constructing} is an interesting multi-hop inference-focused dataset created using entity-relation pairs for queries spanning different Wikipedia passages. While the focus of our pilot study is similar to theirs, we believe that our method can easily be extended to other inference types. Also, identifying the correct span is more challenging than choosing an answer from a list.

We aim to incorporate multiple language styles, making it hard for the system to memorize linguistic patterns \cite{williams2017broad}. We achieve this by using two parallel passages that talk about the same or related subject(s) but are obtained from different sources. This helps in formulating referential inference questions because there exists no single sentence in the passage which matches a paraphrase of the question, and necessitates that inference (which goes beyond co-reference) be performed across both passages. Evaluation is easy and objective because answers are still spans within the passages. Questions can be answered solely on the basis of the information provided in their accompanying passages.

For example, to answer Question 1 in Table \ref{parex}, the system will have to infer from passage 1 that President Kamazu Banda belongs to the MCP and was defeated in the elections. The equivalence of this event and the election in passage 2 must be established, while comprehending that the ``favored challenger'' Bakili Muluzi is the one Banda lost the elections to, and who belonged to the UDP, making it the correct answer.

Given that the information is spatially scattered across the two passages, this method would ensure that the parallel passages have to be understood in combination to answer the question. 
\begin{table*}[th!]
\centering
\begin{tabular}{p{1.0\linewidth}}
\hline
\small
Hastings Kamuzu Banda was the leader of Malawi from 1961 to 1994. In 1963 he was formally appointed prime minister of Nyasaland and, a year later, led the country to independence as Malawi. Two years later he proclaimed Malawi a republic with himself as president. He declared Malawi a one-party state under the Malawi Congress Party (MCP) and became President of MCP as well as President for Life of Malawi in 1971. \textcolor{orange}{A referendum} ended his one-party state and a special assembly ended his life-term presidency, stripping him of most of his powers. Banda ran for president in the democratic elections which followed and was defeated. He died in South Africa in 1997. \\ \hline
\small
Malawians Saturday wound up an historic election campaign bringing multiparty politics to a country ruled for the past three decades by President Hastings Kamuzu Banda. The ailing president inspected troops from an open truck as some 20,000 people turned up at a stadium here to celebrate his official birthday ahead of elections on May 17. Reading a prepared speech with some difficulty, Banda appealed to Malawians \textcolor{blue}{to conduct themselves "as ladies and gentlemen"} during the elections, which should be "free and fair." Meanwhile, the bigger opposition rally was addressed by the presidential challenger favored to win the elections, Bakili Muluzi of the \textcolor{red}{United Democratic Front (UDF)}.\\ \hline
  \small
  \textcolor{red}{Question 1}: Who emerged victorious between the MCP and UDF?\\
  \small
  \textcolor{blue}{Question 2}: What did the MCP leader ask of the people of Malawi on polling day?\\
  \small
  \textcolor{orange}{Question 3}: What brought multiparty politics to Malawi after three decades?\\
 \hline
\end{tabular}
\caption{Example of a Parallel Passage. The questions and corresponding answers are color coded.}
\label{parex}
\end{table*}

\section{Proof of Concept}
For a fair evaluation of existing models, we sought to use data drawn from a similar domain, but written in a different style. 
We chose the CNN/Daily Mail corpus and Wikipedia because they both focus on factoid statements, yet differ in language style to a noticeable extent (e.g. in the use of idiomatic expressions).
We picked 20 CNN/Daily Mail articles at random to form one of the passages in our pair. To find an associated parallel passage, we selected the most frequently mentioned entities in each article and obtained its corresponding Wikipedia pages. We fragmented these into passages with at most 500 words, and performed a k-Nearest Neighbor search using tf-idf and topic vectors \cite{blei2003latent} to form pairs. We tuned the number of entities per article used to retrieve Wikipedia pages, as well as the sections considered in each article. This process produced a total of 15 News-Wiki passage pairs. While no two pairs have the same news article, they may be paired with the same Wiki passage. 

We focused on referential inference for this pilot, but the method can be extended to include questions based on other types of inference. 15 human annotators were given explicit instructions and real-world examples to form question-answer pairs using given parallel passages. We collected $\sim$ 50 valid question-answer pairs through this mechanism. The average length of the answers obtained was around 4 words. Basic sentence retrieval statistics (similar to the ones discussed in Section 2) are shown in Table \ref{statsexpts}, indicating that lexical similarity between the question and passage sentences is insufficient to obtain an answer. Our small-scale experiment shows the feasibility of the approach, although collecting a larger dataset requires more effort in acquiring passages and generating questions from diverse sources.

\section{Analysis of Existing Models}

\begin{table}[ht!]
\centering
\begin{tabular}{|c|c|c|c|c|}
\hline
\multirow{2}{*}{Model} & \multicolumn{2}{c|}{\textbf{SQuAD}} & \multicolumn{2}{c|}{\textbf{ParallelQA}} \\ \cline{2-5} 
                       & EM           & F1          & EM             & F1             \\ \hline
BiDAF &  67.70  &  77.30  &  35.29  &  42.52  \\
DrQA  &  69.64  &  78.76  &  39.22  &  47.23  \\
R-Net &  71.07  &  79.51  &  41.18  &  50.38  \\
\hline
\end{tabular}
\caption{Performance on SQuAD vs ParallelQA}
\label{eval-expt}
\end{table}

\begin{table*}
\small
\centering
\resizebox{\textwidth}{!}{
\begin{tabular}{|p{0.7\linewidth}|p{0.3\linewidth}|}
\hline
\textbf{Passage} & \textbf{Question}  \\ \hline
...\textcolor{red}{The UN} is the largest, most familiar, most internationally represented and most powerful intergovernmental organisation in the world...UN envoy \textcolor{blue}{Yasushi Akashi} called a meeting of all parties to talks on a four-month ceasefire for Saturday afternoon, he added... &    Who was sent to Bosnia as the envoy of most powerful intergovernmental organisation in the world?   \\ \hline
...On arrival, the president and his wife Hillary were taken to \textcolor{red}{University College}, one of 37 Oxford colleges, where he studied political science as a Rhodes Scholar between October 1968 and June 1970... Clinton was born and raised in \textcolor{blue}{Arkansas} and ... &    From which state was this US President who was a Rhodes scholar between 1968 and 1970?  \\ \hline
...withdrew from a UN-designated three-kilometer (two-mile) exclusion zone around the \textcolor{blue}{eastern Bosnian enclave of Gorazde} ... The United Nations (UN) is an intergovernmental organization ... A replacement for the ineffective League of Nations, the organization ... [\textcolor{olive}{eastern Bosnian enclave}, \textcolor{magenta}{Gorazde}, \textcolor{cyan}{eastern Bosnian enclave of Gorazde}] & Where in Bosnia did the successor of the League of Nations designate an exclusion zone?  \\ \hline
\textcolor{blue}{Todd Martin} squeezed to a 7-6 7-6 victory over \textcolor{olive}{fellow-American} \textcolor{red}{Pete Sampras} in the final of the Queen's Club tournament here on Sunday. The win further bolstered fifth seeded Martin's reputation as one of the most dangerous grass court players... & Pistol Pete lost to whom in the Queen's club tournament?  \\ \hline
... (RENAMO) rebels at a UN-supervised assembly point brutally beat one of their senior officials during a mutiny over severance pay on June 1 at \textcolor{blue}{Mocubela}, about 100 kilometers (62 miles) east of Mocuba. But RENAMO has denied the official, identified as Raul Dique, was beaten up by \textcolor{magenta}{mutineers}, the Mozambican news agency (AIM) said in a report monitored in \textcolor{orange}{Harare Thursday} ... & Where was Raul Dique beaten up by rebels of RENAMO? \\ \hline
\end{tabular}
}
\caption{Examples of error trends on ParallelQA: blue - gold answer, red - span predicted incorrectly by all models,  orange - BiDAF and R-Net prediction overlap, olive - BiDAF, magenta - DrQA, cyan - R-Net}
\label{examples-test}
\end{table*}

We consider three deep learning models: Bidirectional Attention Flow (BiDAF) \footnote{\tiny{\url{https://allenai.github.io/bi-att-flow/}}} \cite{seo2016bidirectional}, Document Reader (DrQA) \footnote{\tiny{\url{https://github.com/hitvoice/DrQA}}} \cite{chen2017reading}, and Gated Self-Matching Networks (R-Net) 
\footnote{\tiny{\url{https://github.com/HKUST-KnowComp/R-Net}}} \cite{wang2017gated} trained on SQuAD. We feed the concatenated parallel passage and the question as inputs. On a total of 51 QA pairs, we observed exact match (EM) scores of about 40\% and token overlap F1 scores of about 45\% for all models, versus their performance on the SQuAD dataset (EM of almost 70\% and F1 of 80\%). Detailed results are shown in Table \ref{eval-expt}. 

Although the models were trained and tested on different datasets, we expect them to perform reasonably well on the new task since the data sources and domain are similar. 
Also, the size of our collected data is much smaller than the SQuAD development set, but we believe that the samples are fairly representative of data that can be generated using our proposed mechanism. Thus, the low EM and F1 scores support our hypothesis that these datasets do not adequately assess  the capabilities of these models, which overfit to lexical patterns rather than generalizing.

\begin{figure*}[h!]
\centering
\includegraphics[width=\linewidth]{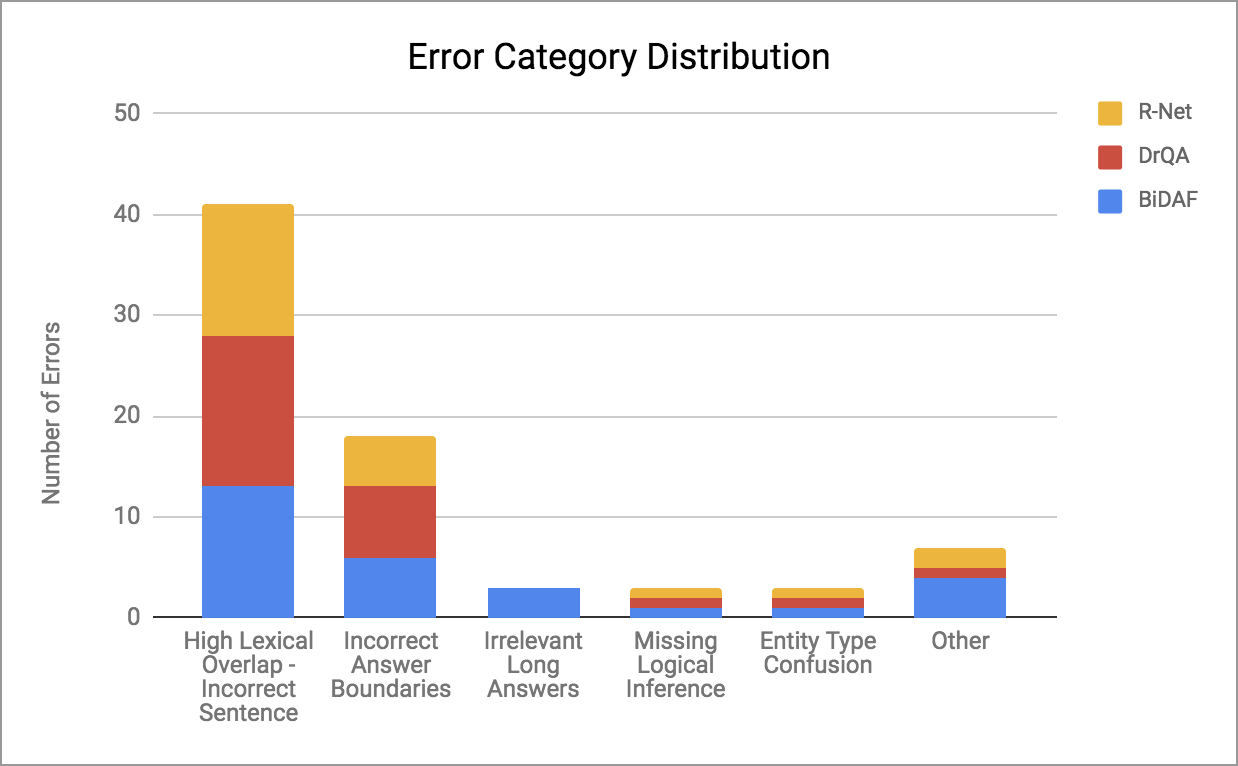}
\caption{Distribution of errors by BiDAF, DrQA and R-Net across different categories using manual inspection}
\label{dist-err}
\end{figure*}

We now discuss a few common errors observed upon manual inspection of the results. Examples for each are provided in Table \ref{examples-test}. The distribution of predictions across these error categories can be found in Figure \ref{dist-err}, Section \ref{sec:supplemental}.
\begin{itemize}
\setlength\itemsep{-0.2em}
\item\textbf{High Lexical Overlap - Incorrect Sentence}: The models tend to pick answer spans from sentences which have high lexical overlap with the question. We observe that this accounts for the largest chunk of errors across all models (example 2). Our observations are consistent with the findings of \citet{jia2017adversarial}. The models often simply resolve the referential expression in the question to its corresponding entity. In example 1, the models resolve ``organisation'' in the question to ``The UN'' due to high lexical similarity.

\item\textbf{Incorrect Answer Boundaries}: This is the second most frequently observed error, where the answers generated are almost correct, but models face issues in appropriately defining answer boundaries (example 3). R-Net and DrQA, on average, produce shorter answers. BiDAF tends to produce longer answers.

\item\textbf{Missing Logical Inference}: Models are sometimes unable to make certain logical conclusions like A's victory over B implies that B lost to A (example 4).

\item\textbf{Entity Type Confusion}: Despite having a variety of entities as answers to questions in the training data, sometimes the model answers do not correspond to the correct entity type (example 5).
\end{itemize}

\section{Discussion \& Conclusion}
While our approach is promising, we observed a few problems during the pilot study. Longer passages and constraints on the question formulation require more time and skill in the annotation process. This can lead to crowd-workers formulating a single referring expression and then using it in different contexts to form questions, reducing diversity. For some questions, although inference is needed, both passages may not be necessary to answer them. Since we used news articles and Wikipedia passages in our pilot study, 58.82\% of answers were named entities. We plan to extend this mechanism to other inference types and conduct a larger pilot before scaling up the collection.

Our experiments demonstrate that the ParallelQA task can be more challenging than some prior QA tasks. Our analysis shows that many popular RC datasets seem to test the ability of models to pick up superficial cues. ParallelQA is our proposed step towards inference-oriented reading comprehension. We use parallel passages from different sources for generating reasoning questions which encourage systems to gain a deeper understanding of language, and become robust to variations in style and topic. We include examples from our initial pilot study in Table \ref{extra-par-ex}. 

\section*{Acknowledgments}
The authors would like to thank Chaitanya Malaviya, Sandeep Subramanian, Siddharth Dalmia, Tejas Nama and Vaishnavi AK for useful discussions. We would also like to express our gratitude to the annotators who participated in the pilot study.


\bibliography{naaclhlt2018}
\bibliographystyle{acl_natbib}
\pagebreak
\appendix
\onecolumn
\section{Supplemental Material}
\label{sec:supplemental}
\begin{table*}[h!]
\begin{center}
\resizebox{\textwidth}{!}{
\begin{tabular}{|c|c|c|c|} 
\hline
\textbf{Inference Type} & \textbf{Meaning} & \textbf{Examples} & \textbf{Information Required} \\ \hline
Referential & Coreferences, Referring Expressions & Bill Clinton's wife is Hillary Clinton & A link between the expression and entity it refers to \\
Figurative & Metaphors & All the world's a stage & A dictionary of common metaphors and what they mean \\
Part-Whole & Inclusion & A dog is an animal & An ontology of hierarchical and other relationships between words \\
Numeric & Units, Operations & 60 seconds is a minute & Equivalence (and conversion) of units, Basic Operation Skills \\
Lexical & Meanings from Linguistic Context & I ate an apple (apple = fruit or company?) & Contextual Information: Word Embeddings / NER / PoS \\
Denotation & Literal Meanings of Expressions & Olive branch denotes peace & World Knowledge + Contextual Information \\
Spatial & Reasoning about Space & Berlin is in Germany which is in Europe & World Knowledge + Basic Spatial Reasoning Rules \\
Temporal & Reasoning about Time & World War II happened before Cold War & World Knowledge + Basic Temporal Reasoning Rules \\
\hline
\end{tabular}
}
\end{center}
\caption{Different Types of Inference along with examples and possible information required to perform them}

\label{inf-types}
\end{table*}

\begin{table*}[th!]
\centering
\begin{tabular}{p{1.0\linewidth}}
\hline
\small
\textcolor{red}{Todd Martin} squeezed to a 7-6 7-6 victory over fellow-American Pete Sampras in the final of the \textcolor{blue}{Queen's Club tournament} here on Sunday. The win further bolstered fifth seeded Martin's reputation as one of the most dangerous grass court players. But it was essentially a baseline slogging match which provided little to whet the appetite for Wimbledon. There were no breaks of serve in either set and only three break points in the entire match - two against Sampras in the second game and one against Martin in the next. Martin clinched the first tie-break courtesy of a double fault from Sampras to lead 4-2 and then a glorious cross-court forehand return on his second set point to take the shoot-out 7-4. He took the second tie-break by the same score, Sampras saving three match points before a fierce smash clinched Martin's \textcolor{orange}{third} career title and his first victory over his compatriot in four meetings. \\ \hline
\small
Petros "Pete" Sampras (born August 12, 1971) is a retired American tennis player widely regarded as one of the greatest in the history of the sport. He was a longtime world No. 1 with a precise serve that earned him the nickname "Pistol Pete". His career began in 1988 and ended at the 2002 US Open, which he won, defeating rival Andre Agassi in the final. Sampras was the first man to win 14 Grand Slam singles titles (seven Wimbledon, five US Open, two Australian Open). He also won seven year-end championships and finished six consecutive seasons atop the rankings. Summary of professional awards. U.S. Olympic Committee "Sportsman of the Year" in 1997. He was the first tennis player to receive this award. GQ Magazine's Individual Athlete Award for Man of the Year in 2000. Selected the No. 1 player (of 25 players) in the past 25 years by a panel of 100 current and past players, journalists, and tournament directors to commemorate the 25th anniversary of the ATP in 1997. Voted 48th athlete of Top 50 Greatest North American Athletes of ESPN's SportsCentury (also youngest on list). In 2005, TENNIS Magazine named Sampras the greatest tennis player for the period 1965 through 2005, from its list, "The 40 Greatest Players of the TENNIS Era".
\\ \hline
  \small
  \textcolor{red}{Question 1}: The first man to win 14 Grand Slam singles titles lost to whom in the Queen's club tournament?
\\
  \small
  \textcolor{blue}{Question 2}: The greatest tennis player for the period 1965 through 2005 lost to Todd Martin in the finals of which tournament?\\
  \small
  \textcolor{orange}{Question 3}: What was the tally of Todd's career titles after defeating the GQ Magazine's Man of the Year award winner, in the final of Queen's club tournament?
\\ \hline 
\\ \hline
\small
Cambodian co-premiers \textcolor{orange}{Prince Norodom Ranariddh} and Hun Sen said Wednesday they had agreed to holding peace talks with the Khmer Rouge in Pyongyang without preconditions, in response to an appeal by King Norodom Sihanouk. The co-premiers had sent an official letter to the king "saying that we are ready to go to Pyongyang without ceasefire, without preconditions," Prince Ranariddh told journalists. "Let talks begin," he added. Hun Sen said the talks, beginning on May 27, would be based on \textcolor{red}{a peace plan put forward by King Sihanouk}, but added that the government had yet to receive a reply from the Khmer Rouge regarding the proposal. King Sihanouk has proposed that certain "acceptable" members of the Khmer Rouge be given senior cabinet posts in the government in exchange for giving up their zones, ceasing all guerrilla activities and merging their fighters with the royal armed forces.\\ \hline
\small
Hun Sen is the Prime Minister of Cambodia, President of the Cambodian People's Party (CPP), and Member of Parliament (MP) for Kandal. He has served as Prime Minister since 1985, making him the longest serving head of government of Cambodia, and one of the longest serving leaders in the world. From 1979 to 1986 and again from 1987 to 1990, Hun Sen served as Cambodia's foreign minister. His full honorary title is Samdech Akeak Moha Sena Padey Techo Hun Sen. Born Hun Bunal, he changed his name to Hun Sen in 1972 two years after joining the Khmer Rouge. Hun Sen rose to the premiership in January 1985 when the one-party National Assembly appointed him to succeed Chan Sy who had died in office in December 1984. He held the position until the 1993 UN-backed elections, which resulted in a hung parliament. After contentious negotiations with the FUNCINPEC, Hun Sen was accepted as Second Prime Minister, serving alongside Norodom Ranariddh until a \textcolor{blue}{1997} coup which toppled the latter. Ung Huot was then selected to succeed Ranariddh. 
\\ \hline
  \small
  \textcolor{red}{Question 1}: According to Hun Bunal, what is the basis of talks on May 27th?
\\
  \small
  \textcolor{blue}{Question 2}: Until which year did the Cambodian co-premiers hold office?\\
  \small
  \textcolor{orange}{Question 3}: The President of the Cambodian People's Party was holding peace talks with the Khmer Rouge along with whom?
\\ \hline 
\end{tabular}
\caption{Examples of collected parallel passages. The questions and corresponding answers are color coded.}
\label{extra-par-ex}
\end{table*}
\end{document}